\documentclass[journal]{IEEEtai}

\usepackage[colorlinks,urlcolor=blue,linkcolor=blue,citecolor=blue]{hyperref}

\usepackage{newtxtext}         
\usepackage{xcolor}            
\usepackage{color,soul}        
\usepackage{color,array}       
\usepackage{ulem}
\usepackage[none]{hyphenat}

\usepackage{amsmath}           
\usepackage{amssymb}           

\usepackage[pdftex]{graphicx}  
\usepackage{caption}           
\usepackage{tabularx}          
\usepackage{multirow}          
\usepackage{nicematrix,tikz}   

\usepackage{algorithm}         
\usepackage{algpseudocode}     
\usepackage{algorithmicx}      
\usepackage{algpseudocode}      
\usepackage{pgffor}
\algdef{SE}[DOWHILE]{Do}{doWhile}{\algorithmicdo}[1]{\algorithmicwhile\ #1}%
\algrenewcommand\algorithmicrequire{\textbf{Input:}}    
\algrenewcommand\algorithmicensure{\textbf{Output:}}    

\usepackage[noadjust]{cite}    

\begin{document}

\title{DDD-GenDT: Dynamic Data-driven Generative Digital Twin Framework}

\author{\IEEEauthorblockN{Yu-Zheng Lin\IEEEauthorrefmark{2}, Qinxuan Shi\IEEEauthorrefmark{3}, Zhanglong Yang\IEEEauthorrefmark{3}, Banafsheh Saber Latibari\IEEEauthorrefmark{2}, 
Shalaka Satam\IEEEauthorrefmark{2},\\ Sicong Shao\IEEEauthorrefmark{3}, Soheil Salehi\IEEEauthorrefmark{2} (Member, IEEE), and Pratik Satam\IEEEauthorrefmark{1}}
\\
\IEEEauthorblockA{\IEEEauthorrefmark{2}Department of Electrical and Computer Engineering, University of Arizona\\
\IEEEauthorrefmark{1}Department of Systems and Industrial Engineering, University of Arizona\\
\IEEEauthorrefmark{3}School of Electrical Engineering and Computer Science, University of North Dakota\\
Email: \{\IEEEauthorrefmark{2}yuzhenglin, \IEEEauthorrefmark{1}pratiksatam, \IEEEauthorrefmark{2}banafsheh, \IEEEauthorrefmark{2}shalakasatam, \IEEEauthorrefmark{2}ssalehi\}@arizona.edu; \\
\{\IEEEauthorrefmark{3}qinxuan.shi, \IEEEauthorrefmark{3}zhanglong.yang, \IEEEauthorrefmark{3}sicong.shao\}@und.edu
}
\thanks{}}

\maketitle

\begin{abstract}
Digital twin (DT) technology enables real-time simulation, prediction, and optimization of physical systems, but its practical deployment often faces challenges related to high data requirements, proprietary data constraints, and limited adaptability to evolving system conditions. This work introduces DDD-GenDT, a dynamic data-driven generative digital twin framework grounded in the Dynamic Data-Driven Application Systems (DDDAS) paradigm. The proposed architecture comprises of the Physical Twin Observation Graph (PTOG) for representing operational states of the physical twin (PT), an Observation Window Extraction process for capturing relevant temporal state sequences, a Data Preprocessing Pipeline within an LLM-Based Behavior Prediction Engine for sensor data structuring and filtering, and an LLM ensemble that performs zero-shot predictive inference. By leveraging generative artificial intelligence, DDD-GenDT reduces the need for extensive historical datasets, enabling DT construction in data-scarce environments while maintaining privacy for proprietary industrial processes. The DDDAS-driven feedback mechanism allows the DT to autonomically adapt its predictive behavior in alignment with PT-specific wear and degradation patterns, thereby supporting DT-aging, which is the progressive synchronization of the DT with the evolving physical system. The proposed framework is validated using the NASA CNC milling dataset, with spindle motor current as the monitored variable. In a zero-shot prediction setting, the GPT-4-based DT achieves an average RMSE of 0.479 A (4.79\% of the maximum 10 A spindle current), accurately modeling both nonlinear process dynamics and changes arising from PT aging without retraining. These results demonstrate that DDD-GenDT provides a generalizable, data-efficient, and adaptive DT modeling approach, bridging generative AI capabilities with the performance and reliability requirements of industrial DT applications.
\end{abstract}

\begin{IEEEImpStatement}

This paper advances the development of high-fidelity digital twins (DTs) by introducing a reference architecture that formalizes the interaction between physical and virtual spaces and defines essential DT performance metrics. It presents the Dynamic Data-Driven Generative Digital Twin (DDD-GenDT) architecture, which integrates a Physical Twin Observation Graph (PTOG), observation window extraction, and a large language model-based prediction pipeline to enable zero-shot future state estimation. By aligning with the Dynamic Data-Driven Application Systems (DDDAS) paradigm, the architecture supports adaptive updates at process epoch points, allowing the DT to autonomically age alongside the physical system. The use of generative AI eliminates the need for training datasets, addressing the cost constraints of data collection in Industry 4.0 environments. Experimental validation using a NASA CNC milling dataset demonstrates the framework’s ability to predict physical twin behavior with high accuracy and adapt dynamically to tool wear, thereby confirming its effectiveness in low-data, real-time industrial scenarios.
\end{IEEEImpStatement}


\section{Introduction}
Recent advances in computing, communication, and artificial intelligence (AI) are enabling the design and deployment of intelligent systems that integrate sensing, connectivity, and computational intelligence into industrial processes. These developments are driving the transformation of traditional manufacturing into smart manufacturing systems under the paradigm of Industry 4.0 (I4.0)~\cite{min2019machine, kaur2020convergence, rathore2021role}. The core vision of I4.0 lies in achieving cyber–physical integration, where physical processes are tightly coupled with their digital representations to enhance operational efficiency, resilience, and adaptability. The networked nature of I4.0 systems, supported by ubiquitous sensing and cloud-based data management, has enabled the emergence of Digital Twin (DT) technology. DTs serve as computational counterparts of physical systems, continuously synchronized with their real-world equivalents through bidirectional data flows~\cite{rao2023developing, cimino2019review}. This capability allows DTs to monitor, simulate, and predict the behavior of physical assets across their entire life cycle, enabling informed decision-making in real time.

A DT is a software system that models the behavior of its corresponding Physical Twin (PT), evolving in parallel with it by autonomously updating its internal state to reflect the PT’s operational condition and gradual changes due to wear, environmental effects, and usage patterns. This self-adaptation, or ``DT-aging,'' ensures that the digital model remains representative of the specific PT instance over time. Unlike static models, a well-designed DT is dynamic, context-aware, and capable of maintaining behavioral fidelity despite environmental perturbations and component degradation~\cite{kaur2020convergence, rathore2021role}. The applications of DTs are wide-ranging, encompassing sectors such as predictive maintenance in manufacturing, real-time operational optimization in energy systems, personalized treatment in healthcare, and advanced climate modeling for environmental management~\cite{coorey2021health, rao2023developing, cimino2019review}. In manufacturing, DTs enable optimization of production lines, early detection of defects, and proactive maintenance scheduling, thereby reducing downtime and improving product quality. In healthcare, DTs facilitate patient-specific simulations for treatment planning, while in infrastructure systems they enhance predictive monitoring and fault mitigation.

Despite their promise, the deployment of DTs in I4.0 environments faces two major challenges. The first challenge is data scarcity. In manufacturing contexts, data collection is inherently tied to physical production: acquiring 1{,}000 data points may require manufacturing 1{,}000 units, each consuming material and operational resources. Furthermore, industrial data is often proprietary, with manufacturers reluctant to share datasets due to concerns over intellectual property, potential counterfeiting, and competitive disadvantage. This scarcity hinders the development of high-fidelity DT models, particularly in complex systems that require accurate modeling of multiple interdependent subsystems. The second challenge is PT-aging, the process by which a physical asset’s characteristics change due to cumulative wear, material degradation, and operational history. PT-aging is deployment-specific, meaning that two identical machines operating in different environments or under different workloads may age differently. A DT must therefore autonomously evolve in a manner that mirrors the unique aging trajectory of its specific PT. This adaptation ensures that predictive analytics, control strategies, and optimization routines remain valid throughout the PT’s operational lifespan. Failure to address PT-aging leads to divergence between the DT’s predictions and the PT’s actual behavior, reducing the system’s reliability and decision-making accuracy.

To address these challenges, this paper proposes the Dynamic Data-Driven Generative Digital Twin (DDD-GenDT) architecture, which unifies a Physical Twin Observation Graph (PTOG), an observation window extraction process, and an LLM-based Behavior Prediction Engine into a cohesive DT development framework. The PTOG captures the PT’s operational state space and the contextual relationships between measurements, serving as a structural foundation for state representation. The observation window extraction process segments time-series data into temporally relevant intervals that preserve dynamic patterns for prediction. The LLM-based Behavior Prediction Engine then processes these observation windows through a two-stage pipeline: a preprocessing module that selects relevant historical observations to construct structured prompts, and a prediction module that employs an ensemble of Large Language Models (LLMs) to produce statistical point estimates of future states. The proposed architecture operates in a zero-shot setting, eliminating the need for task-specific training datasets~\cite{gruver2024large}. This property is particularly advantageous for I4.0 applications where data acquisition is expensive or constrained. By integrating with the Dynamic Data-Driven Application Systems (DDDAS) paradigm~\cite{darema2023dynamic}, the DDD-GenDT supports autonomous DT-aging, updating its behavior at defined process epochs to reflect the evolving condition of its PT. This combination of zero-shot prediction and adaptive aging brings the DT closer to the concept of a ``true'' twin, capable of both accurate future state estimation and self-evolution in parallel with its physical counterpart.

The effectiveness of the proposed approach is demonstrated through a CNC machining case study using the NASA milling dataset. In this scenario, the DDD-GenDT framework builds two separate DTs to predict spindle motor current behavior during machining operations. The experimental results confirm that the DTs can predict future PT behavior with high accuracy despite the absence of prior training data, and can dynamically adapt to changes caused by tool wear over successive machining runs. This validation underscores the architecture’s potential for delivering high-fidelity, adaptive DTs in real-world industrial environments.

The main contributions of this paper are as follows:
\begin{itemize}
    \item The paper introduces a reference architecture for digital twins to guide development and implementation, emphasizing the interaction between physical and virtual spaces and defining essential DT performance metrics.
    
    \item The paper presents the Dynamic Data-Driven Generative Digital Twin (DDD-GenDT) architecture, integrating a Physical Twin Observation Graph (PTOG), observation window extraction, and an LLM-based prediction unit for future state estimation.
    
    \item The paper connects the DDD-GenDT framework to the Dynamic Data-Driven Application Systems (DDDAS) paradigm, enabling adaptive model updates at process epoch points based on prior run measurements, enabling the DT to age autonomously as the PT ages due to wear and tear.
    
    \item The presented DDD-GenDT architecture explores the use of generative AI to model and predict the PTs future behavior. This is achieved via a zero-shot capability, requiring no training datasets; making the framework ideally suited for Industry 4.0 systems where there is cost associated with data collection. The zero-shot learning capability, in combination with the DT aging capability, makes the proposed DDD-GenDT behavior close to a 'true' DT.
    
    \item The paper validates the proposed DDD-GenDT approach on a CNC machining scenario, by building two separate DTs from the NASA CNC milling dataset. The experimental results demonstrate the accuracy of the proposed DDD-GenDT based DTs, in predicting future PT behavior with no prior training. The experimental results also demonstrate the DTs ability to modify and age itself as the CNC machine goes through tool-wear, a result of multiple cutting runs; thus, demonstrating high-fidelity predictions and dynamic adaptation to PT-aging using minimal data.
\end{itemize}

The rest of the paper is organized as follows: In Section \ref{sec:background}, we review studies of digital twin, DDDAS, and LLM application on the physical systems; in Section \ref{sec:ref_architecture}, we propose a digital twin reference model; in Section \ref{sec:main_framework}, we present the DDD-GenDT framework; in Section \ref{sec:exp}, we present the experimental evaluation of the DDD-GenDT framework with the NASA milling dataset, and in Section \ref{sec:conclusion}, we conclude the paper. 

\section{Background}\label{sec:background}
\subsection{Digital Twin}
A Digital Twin (DT) is defined as “software systems replicating the behavior of one or more physical processes using one or more behavior models” and aims to establish a mirrored connection between the physical and virtual realms by mapping sensor-measured data onto the virtual model \cite{lin2023dt4i4}. However, a DT is not merely a model. It is a software system that represents the complete lifecycle of its physical twin, covering the design phase, operational usage, and the effects of wear and tear over time \cite{lin2025respond}. In this work, this gradual evolution of the digital twin in response to the physical twin’s degradation and operational history is referred to as DT-aging. In practice, the virtual representation interacts with the physical system through continuous data exchange, enabling bidirectional influence: the DT reflects the physical state and provides feedback for adaptive control based on its outputs. This concept originated in aerospace engineering, where Shafto et al. introduced the term in the 2010 NASA Technology \& Processing Roadmap draft \cite{shafto2010draft}, defining it as simulation-based systems engineering. The report emphasized that high-fidelity DTs integrate multiphysics and multiscale simulations with sensor updates and historical data, enabling the continuous prediction of system behavior. Such predictions facilitate the assessment of vehicle parameters, abnormal conditions, and environmental effects, thereby improving mission success rates. Despite these advantages, practical deployment is constrained by computational resources and modeling limitations; constructing complex, high-fidelity behavioral physics models remains challenging, and in some cases, the resulting simulations cannot fully capture or interact with the physical behavior \cite{li2024machine}.

 With the maturation of artificial intelligence (AI) and the advancement of computer computing power, digital twin modeling based on machine learning has presented new opportunities for developing digital twins \cite{kaur2020convergence, rathore2021role, zhang2023lifetime}. It enables the modeling of complex systems that previously required substantial computing resources and were challenging to formulate using physics and statistics. The machine learning-based DT developments demonstrate the great potential of machine learning applications in building digital twins. More insights into the physical process can be obtained through the continuous interaction between the machine learning-based digital twin model and the physical system. Recent works further illustrate this trend by integrating deep learning techniques with multitask objectives. For instance,  Zhang et al. \mbox{\cite{zhang2025process}} proposed a multitasking deep learning framework for the process monitoring of tower pumping units operating under variable conditions. Their approach jointly performs fault type classification and regression of operating indicators using a shared encoder and task-specific decoders. Experimental results on real-world industrial data demonstrate the superiority of the multitask model over conventional single-task and statistical baselines. This highlights the increasing emphasis on scalability and interpretability in ML-based digital twins for complex industrial systems.

\subsection{Dynamic Data Driven Applications Systems}

Dynamic Data-Driven Application Systems (DDDAS) is a computational paradigm in which computational and instrumentation aspects of an application are dynamically integrated in a feedback control loop, allowing dynamic integration of sensed observations into the application model while allowing the application model to control the instrumentation \mbox{\cite{rasheed2020digital, darema2023dynamic}}. From an architectural perspective, DDDAS encompasses three foundational elements: (1) real-time data assimilation, enabling continuous adjustment of model states based on streaming physical measurements; (2) instrumentation control, which dynamically directs data acquisition to maximize informativeness; and (3) computational integration across heterogeneous platforms, ranging from edge devices to high-performance computing infrastructures, making DDDAS architecture well suited for operational decision-making in complex, multiscale, and data-intensive environments. Papadimitriou et al. \mbox{\cite{papadimitriou2023dynamic}} successfully apply the DDDAS framework to predict adverse weather like tornadic events, their onset paths, and their impact, evaluating multi modal, space, and time distributed data from heterogeneous sources like the Next Generation Weather Radar (NEXRAD), in conjunction with phased radar array instrumentation, as well as temperature, pressure, wind velocity and other related data streams. Papadimitrious' DDDAS based tornado prediction pipeline, consuming large, high speed data streams from different weather sources, deployed on cloud infrastructure like the National Science Foundation (NSF) supported Chameleon, achieved high performance accuracy. These characteristics of the DDDAS architecture, especially its ability to continuously ingest data from instrumentation, update behavior models, and in turn interact with the physical process, make it an inspiration for our proposed DDD-GenDT architecture. We envision the proposed DDD-GenDT architecture, to imbibe all the three foundational elements of the DDDAS architecture, allowing building of DTs that 1) at real-time assimilate to the current working state of the PT, 2) use LLM-Based Behavior Predictions to predict the future behavior of the PT, and 3) integrate back with the PT, to allow for secure and resilient operations. The novel use of the proposed LLM-Based Behavior Predictions in the DDD-GenDT architecture, allows DT design and building in data-scare operations like I4.0 systems, while reducing the deployment and operational costs via non-reliance on expensive cloud infrastructure.   

\subsection{Generative AI and Large Language Model based Behavior Prediction}
Generative Artificial Intelligence (GenAI) are a class of AI techniques, that use deep learning models to produce data that closely resembles real-world distributions \cite{feuerriegel2024generative}, using techniques like Generative Adversarial Networks (GANs), Variational Autoencoders (VAEs), and Transformer-based models enabling the synthesis of high-dimensional inputs \cite{vaswani2017attention}, including images, text, and structured data. These models learn the underlying probability distributions through unsupervised or self-supervised training and are widely applied in simulation, data enhancement, and predictive modeling. In this review, we focus on LLMs, which are transformer-based architectures trained on massive corpora to predict sequences of tokens given prior context. LLMs learn statistical patterns in language that can be adapted to various downstream tasks, including reasoning, code generation, and interaction with physical systems \cite{kasneci2023chatgpt}. Their ability to generalize from few-shot or zero-shot prompts makes them particularly attractive for applications where labeled data is scarce \cite{kojima2022large}. Recent studies have extended LLM usage beyond textual tasks, demonstrating their adaptability for time series forecasting, robotic control, and integration into digital twin systems.

Boiko et al.~\cite{boiko2023autonomous} proposed an LLM-powered system called Coscientist, to automate chemical experiments by controlling heating, stirring, and liquid handling devices, design of GPT-4 based \texttt{Planner Module}, that assists the users through document search, Python code development and experimentation, and research documentation through prompt engineering. The Coscientist demonstrates a measured improvement in research performance across a set of six diverse tasks, while exhibiting potential applications of LLMs in experimental protocol design and semi-autonomous execution. In industrial applications, Gkournelos et al.~\cite{gkournelos2024llm} developed a dual-LLM framework to control collaborative robots using natural language commands, enabling intuitive human-robot interaction on the factory floor. Similarly, Zhao et al.~\cite{zhao2024large} applied LLM-based multi-agent systems to schedule manufacturing tasks dynamically, demonstrating improved flexibility and responsiveness in smart manufacturing environments. These examples highlight the use of LLMs as reasoning agents capable of coordinating physical actions via interface modules and prompt-based behavior programming.

Beyond physical control, researchers have explored the utility of LLMs for modeling time series data. Gruver et al.~\cite{gruver2024large} showed that pretrained LLMs can predict future values in a time series using prompt engineering and statistical estimation, without any fine-tuning. Jin et al. proposed reprogramming time series in natural language form for time series prediction tasks without adjusting LLMs \cite{jin2023time}. Underlying these approaches is the view that LLMs process time series data by treating them as token sequences, where each time step corresponds to a token or group of tokens. During inference, the model autoregressively predicts the next token based on previous ones, which is conceptually similar to forecasting the next time point in a time series.  Recent analysis by Zhou et al.~\cite{zhou2024can} formalized the time series inference process of LLMs as an implicit generative function $G(x_t \mid x_{t-1}, \ldots, x_{t-n})$, and highlighted that LLMs use statistical abstraction rather than explicit temporal modeling. Meanwhile, Garza et al.~\cite{garza2023timegpt} proposed TimeGPT-1, a foundation model trained on extensive time series data using transformer-based architectures. This approach improves scalability and generalization in time series prediction, but typically requires significant training resources. These developments collectively underscore the growing feasibility of using LLMs for sequential prediction tasks. Building on this progress, some researchers have extended LLM applications to the challenging domain of chaotic system prediction. Zhang and Gilpin \cite{zhang2025zeroshot} conducted the first large-scale benchmark of such models on chaotic dynamical systems, a setting characterized by sensitive dependence on initial conditions and predictability limits of approximately one Lyapunov time. Using 135 systems under partial observability, they demonstrated that large foundation models, such as Chronos, achieve short-term accuracy comparable to top fully-trained models and preserve long-term attractor invariants, including fractal dimension and statistical distributions. Their analysis showed performance improvements with increased model size and extended context length, and identified that much of the forecasting ability arises from reusing repeated patterns in the context rather than from explicitly modeling the governing dynamical equations. Despite these promising results, the approach exhibits fragility to shifts in initial conditions, sensitivity to disturbances and nonstationarity, and limited capability for genuine flow-field reconstruction.

However, the integration of LLMs into real-time cyber-physical contexts remains nascent. Our proposed DDD-GenDT framework presents a novel DT architecture that relies on LLMs to predict the PT's future behavior through state space analysis via a spatial and temporal sliding window, while relying on minimal training data and adaptation to the aging of the PT.

\section{The Digital Twin Reference Architecture}\label{sec:ref_architecture}
\begin{figure}[t!]
\centering
\includegraphics[width=1\linewidth]{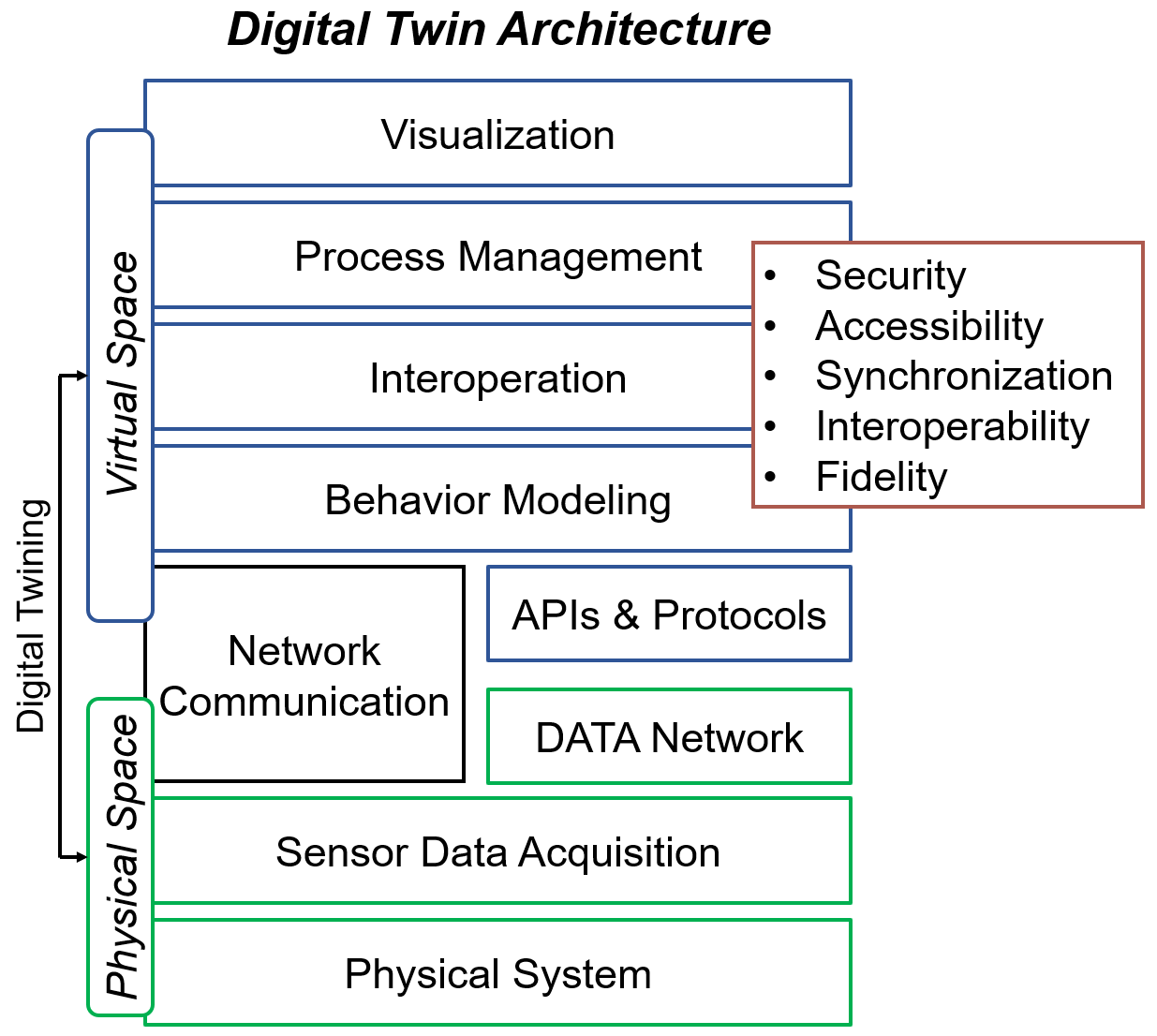}
\caption{The Digital Twin Reference Architecture.}
\label{fig:dt_architecture}
\end{figure} 

\subsection{Digital Twin Reference Architecture}

To address the increasing complexity, heterogeneity, and scale of DT systems, a reference architecture is defined as a foundational guide for their systematic design, implementation, and deployment \cite{piroumian2021digital}. As illustrated in Figure \ref{fig:dt_architecture}, the architecture formalizes the coupling between the \textit{Physical Space} and the \textit{Virtual Space} through a continuous integration and feedback cycle, herein referred to as \textit{digital twinning}. This reference model separates the digital twinning process into two distinct but interconnected domains. The \textit{Virtual Space} encompasses the computational and analytical components of the DT, while the \textit{Physical Space} contains the real-world assets, processes, and sensing infrastructure that generate observational data and receive control inputs. The communication bridge between these spaces is supported by a combination of data networks, communication protocols, and application programming interfaces (APIs), enabling reliable and secure bidirectional information exchange. The architecture is designed to be domain-agnostic, allowing adaptation to a wide range of application-specific requirements in manufacturing, infrastructure, energy, and other cyber–physical domains.

\subsubsection{Virtual Space}
The Virtual Space contains the functional layers responsible for modeling, management, interoperability, and visualization of the DT.
\begin{itemize}
    \item \textbf{Behavior Modeling (DT):} The behavior modeling layer maintains an executable representation of the physical twin (PT) dynamics. It integrates time-stamped observations, control signals, and environmental conditions to infer the PT’s latent state and predict its evolution. This layer enables simulation, forecasting, and what-if scenario analysis. Modeling approaches can include physics-based models, data-driven models, or hybrid combinations, depending on the fidelity requirements and available data.
    \item \textbf{Interoperation:} This layer ensures semantic, syntactic, and temporal compatibility among heterogeneous DT components and with external systems. Interoperation functions include schema harmonization, unit and coordinate transformations, time-base synchronization, and interfacing between models built using different methodologies.
    \item \textbf{Process Management:} The process management layer orchestrates DT workflows, such as data assimilation, model retraining, and software deployment. It manages resource allocation, version control, and execution scheduling to maintain operational consistency with the PT.
    \item \textbf{Visualization:} The visualization layer provides interactive and context-aware dashboards for rendering PT telemetry, DT states, residuals, predictions, and uncertainties. It incorporates role-based access control and information filtering to support decision-making across multiple stakeholders.
    \item \textbf{APIs and Protocols:} This layer defines the control and data exchange interfaces for Virtual Space services. Control interfaces may be implemented using REST or gRPC, while telemetry and event-driven data streams can employ protocols such as OPC UA or MQTT.
\end{itemize}

\begin{figure*}[t!]
\includegraphics[width=\textwidth]{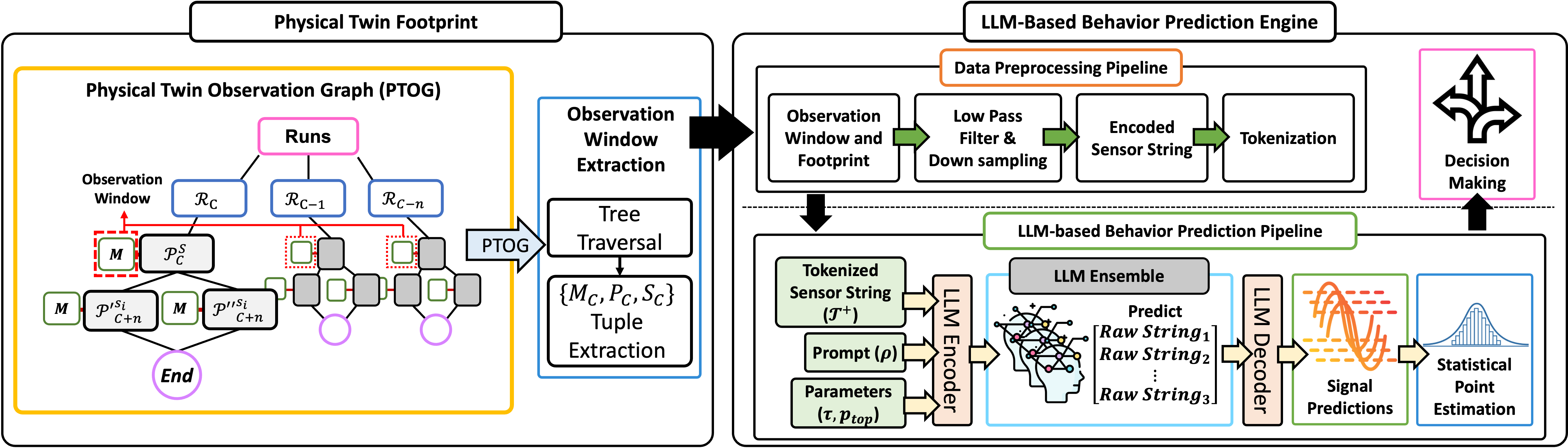}
\caption{Proposed Dynamic Data-Driven Generative Digital Twin (DDD-GenDT) Architecture. The framework integrates the Physical Twin Observation Graph (PTOG) for structured representation of process runs, states, and sensor data with an LLM-based prediction engine. Observation windows extracted from the PTOG are preprocessed, tokenized, and used by an LLM-based behavior prediction engine to forecast future physical states and sensor signals. These predictions support decision-making for adaptive control, enabling synchronization between digital and physical twins in alignment with the DDDAS paradigm.}
\label{fig:Concept_Main}
\end{figure*} 

\subsubsection{Physical Space}
The Physical Space includes the physical assets and associated data acquisition infrastructure.
\begin{itemize}
    \item \textbf{Physical System (PT):} The PT represents the real-world asset, process, or environment being monitored and controlled. It serves as the ground truth reference for evaluating DT accuracy, stability, and synchronization performance.
    \item \textbf{Sensor Data Acquisition:} This layer captures physical quantities of interest using embedded or external sensors. Data collection may be accompanied by calibration, filtering, and edge-level preprocessing to meet latency, reliability, and bandwidth constraints \cite{chuang2019real,karnik2024constrained}.
    \item \textbf{Data Network:} The data network provides the transport layer for both measurement and control signals. It can be implemented using wired industrial Ethernet, wireless networks such as 5G, or hybrid solutions. It must satisfy quality-of-service (QoS) guarantees, redundancy requirements, and temporal synchronization constraints.
    \item \textbf{Network Communication:} This layer acts as the interface between the Physical and Virtual Spaces. It handles protocol translation, data buffering, authentication, and rate control to ensure secure and lossless data transfer.
\end{itemize}

\subsubsection{System-Level Indicators}
To ensure that the DT system remains reliable, secure, and accurate, the architecture incorporates five key system-level indicators:
\begin{itemize}
    \item \textbf{Security:} Protection of measurement data, computational models, and control channels using authentication, encryption, and intrusion detection or prevention systems, while also safeguarding personnel safety and equipment safety through real-time hazard detection, access control, and safe shutdown mechanisms \cite{banerjee2011ensuring,alcaraz2022digital, alipour2015wireless}.
    \item \textbf{Accessibility:} Ensuring that authorized users have reliable access to DT services and data. Accessibility is measured using metrics such as uptime percentage, mean time to recovery, and the enforcement of role-based access policies \cite{mamun4643053federated, satam2022ccri}.
    \item \textbf{Synchronization:} Maintaining temporal and state alignment between the PT and the DT. Synchronization performance is quantified using latency measurements, clock offset values, and state estimation errors \cite{jia2020digital}.
    \item \textbf{Interoperability:} The ability to integrate across heterogeneous devices, platforms, and analytical models. Interoperability is achieved through adherence to shared semantic definitions, open standards, and widely adopted communication protocols \cite{klar2023digital}.
    \item \textbf{Fidelity:} The degree of agreement between DT predictions and observed PT behavior across operational regimes. Fidelity is evaluated using accuracy scores, residual analysis, and uncertainty quantification \cite{zhang2022improved}.
\end{itemize}

\subsubsection{Role of the Reference Architecture}
This reference architecture establishes a structured methodology for designing and deploying DT systems. By explicitly defining the responsibilities of the Physical and Virtual Space components it enables the engineering of scalable and adaptable DT implementations. The architecture supports continuous integration of observational data and control feedback, promotes interoperability through standardized communication, and ensures that system-level indicators can be consistently monitored. These properties make the reference architecture suitable for diverse industrial and scientific applications, while also providing a foundation for specialized extensions, such as the propose Dynamic Data-Driven Generative Digital Twin (DDD-GenDT) architecture described in later sections.

\section{Dynamic Data-driven Generative Digital Twin (DDD-GenDT) Framework}\label{sec:main_framework}

\begin{figure}[t!]
\includegraphics[width=\columnwidth]{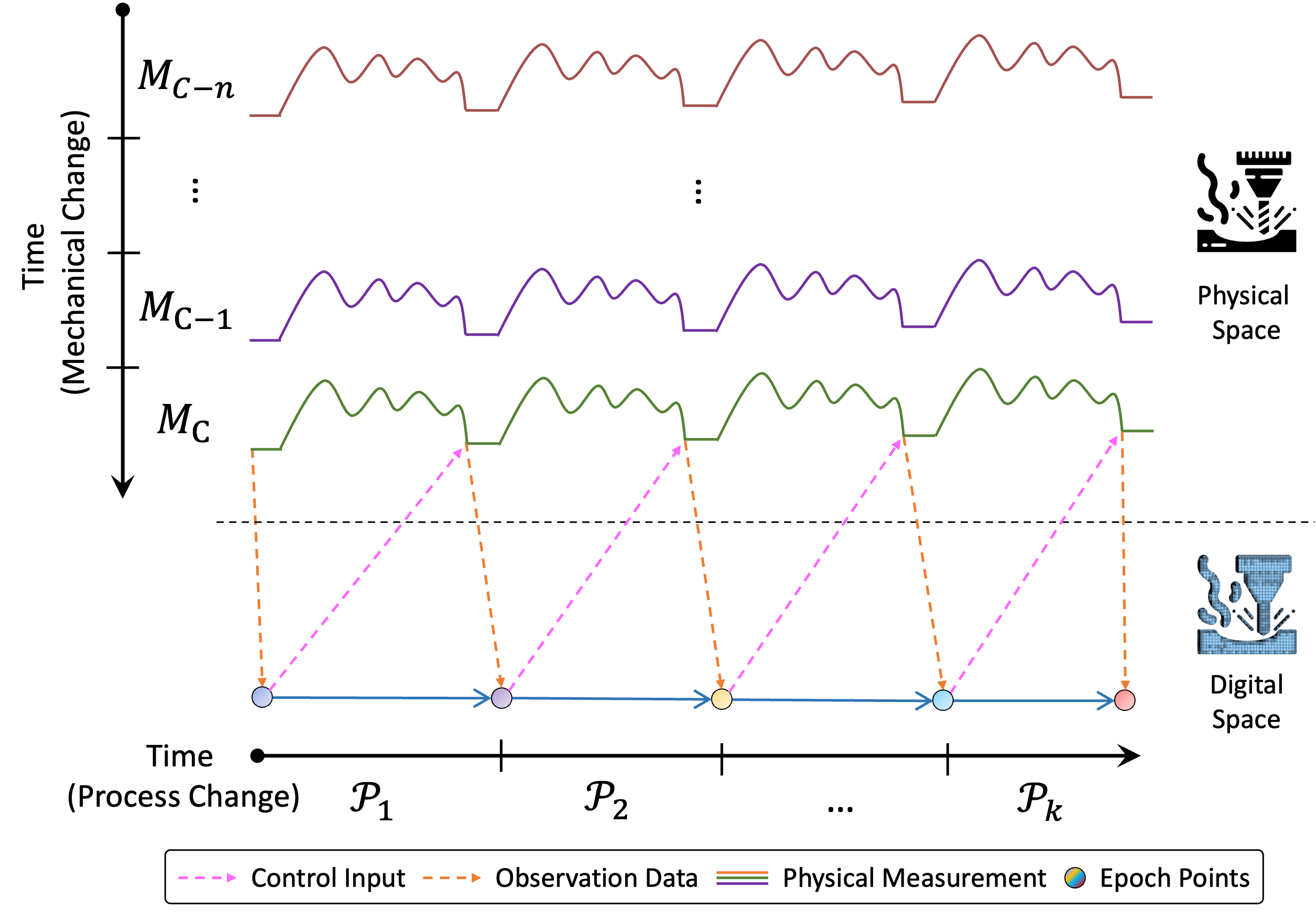}
\caption{Linking DDD-GenDT Architecture to the DDDAS Framework. Vertical axis: mechanical changes over time; Horizontal axis: operation process changes over time. The colored circle represent as epoch points of DT that interactive with physical system.}
\label{fig:Concept_DDDAS}
\end{figure}

This section describes the proposed DDD-GenDT framework as shown in Figure \ref{fig:Concept_Main} and links it with the DDDAS framework as shown in Figure \ref{fig:Concept_DDDAS}.

\subsection{Physical Twin Footprint}

The Physical Twin Observation Graph (PTOG) serves as a structural representation of the temporal and hierarchical organization of data captured from the PT, wherein the PT manufactures multiple product batches, referred to as runs. In the PTOG, each run corresponds to a root-level node, which is subsequently connected to a sequence of process states \( P_c \). These process states constitute the first layer of operational granularity within a run and describe distinct stages of the production workflow.  

Each process state \( P^{s_i}_c \) is monitored through sensor \( s_i \) to produce measurements \( M \) that quantify the PT performance, such as power consumption, temperature, vibration, or force. The PTOG thus encodes both discrete transitions between process states and the continuous acquisition of sensor data associated with each state. This organization enables a unified view of the production execution history, where runs and their constituent process states are systematically linked to their corresponding measurement streams.  

The observation window extraction process operates directly on the PTOG. Its objective is to generate temporally localized segments of historical sensor data that can be used for subsequent modeling and prediction tasks of the DT. This process begins with a traversal of the PTOG that respects both the run-level hierarchy and the sequential ordering of process states within each run. 
The algorithm then moves backward in time across both the current run and previous runs, collecting measurements from the same process state and sensor. This backward search continues until either the observation window reaches the predefined size \( L_w \) or there are no earlier runs containing the process state.

By sliding this observation window along the temporal dimension of the PTOG, the system systematically captures overlapping sequences of sensor readings. Each sequence is represented as a tuple \(\{ M_C, P_C, S_C \}\), where \( M_C \) denotes the measurement vector corresponding to the current window, \( P_C \) specifies the active process state, and \( s_i \) identifies the associated sensor. This tuple representation maintains a consistent mapping between the physical state of the system, the sensing modality, and the numerical data stream.  

The extracted observation windows form the fundamental input units for downstream data preprocessing and predictive modeling. The traversal-based extraction ensures that temporal dependencies and process-state relationships are preserved, while the tuple structure supports unambiguous alignment between physical events and their sensor-based representations. As a result, the PTOG, combined with the observation window extraction process, provides a comprehensive and structured data foundation for digital twin behavior modeling in complex industrial systems.

\subsection{LLM-Based Behavior Prediction Engine}

The LLM-based behavior prediction engine is responsible for forecasting the future state of the physical process and assessing its operational health. The engine comprises two pipelines: the \textit{Data Preprocessing Pipeline} and the \textit{LLM-Based Behavior Prediction Pipeline}. The first pipeline transforms the observation window data into a format suitable for large language model (LLM) inference. The second pipeline uses the processed data to generate state forecasts through an ensemble of LLMs and performs statistical estimation to produce the final prediction for decision-making.

\begin{algorithm}[t]
\caption{Observation Window Extraction from Physical Twin Observation Graph (PTOG)}
\label{alg:observation_window_extraction}
\begin{algorithmic}[1]
\Require PTOG $\mathcal{G} = (\mathcal{R}, \mathcal{P}, \mathcal{S}, \mathcal{M})$, observation window size $L_w$
\Ensure Set of observation windows $\mathcal{W} = \{\{M_C, P_C, S_C\}\}$

\State $\mathcal{W} \gets \emptyset$ \Comment{Initialize observation window set}
\For{$R_c$ in $\mathcal{R}$ in chronological order \textbf{and} $|R_c| < L_w$}
    \For{$P_C$ in $R_c$}
        \State $S_C \gets$ sensor assigned to $P_C$
        \State $M_{\text{hist}} \gets [\,]$ \Comment{Empty measurement list}
        \State $R_{\text{ptr}} \gets R_c$
        \While{$|M_{\text{hist}}| > 0$ \textbf{and} $R_{\text{ptr}} \neq \varnothing$}
            \If{$P_C$ exists in $R_{\text{ptr}}$}
                \State $M_{\text{hist}} \gets M_{\text{hist}} \,\|\, \text{measurement}(S_C, P_C, R_{\text{ptr}})$
            \EndIf
            \State $R_{\text{ptr}} \gets$ previous run in $\mathcal{R}$
        \EndWhile
        \State $M_C \gets \text{concatenate}(M_{\text{hist}})$
        \State $\mathcal{W} \gets \mathcal{W} \cup \{\{M_C, P_C, S_C\}\}$
    \EndFor
\EndFor
\State \Return $\mathcal{W}$
\end{algorithmic}
\end{algorithm}

\subsubsection{Data Preprocessing Pipeline}

The data preprocessing pipeline receives the observation window and the physical footprint generated by the PTOG traversal. This pipeline performs three major operations. First, the observation window and associated footprint are passed through a low-pass filter combined with downsampling. The purpose of this step is to remove high-frequency noise from the sensor data and reduce the sampling density, which improves model efficiency without compromising the representation of key process dynamics.

Second, the filtered and downsampled measurements are encoded into structured sensor strings. Each sensor string represents the temporal sequence of readings from a specific observation window, preserving both the order of measurements and the context of the physical process state. This encoding enables the data to be represented as discrete sequences suitable for token-based modeling.

Third, the encoded sensor strings are passed through a tokenization step. This process maps the raw numerical and symbolic elements of the sensor strings into discrete tokens from a predefined vocabulary. Tokenization ensures that the LLM can process the sensor data using the same mechanisms employed for text-based inputs. The result of this pipeline is a tokenized sensor string $\mathcal{T}^+$ that retains the temporal and structural characteristics of the original measurements while conforming to the input requirements of the prediction engine.

\begin{algorithm}[t]
\caption{Data Preprocessing Pipeline}
\label{alg:data_preprocessing_pipeline}
\begin{algorithmic}[1]
\Require Observation windows $\mathcal{W}=\{\{M_C,P_C,S_C\}\}$ from PTOG; low-pass cutoff $f_c$; downsampling factor $d$; encoder $\mathsf{Enc}(\cdot)$; tokenizer $\Pi(\cdot)$
\Ensure Tokenized dataset $\mathcal{T}^+=\{(\mathbf{t}_j,P_C,S_C)\}$

\State $\mathcal{T}^+ \gets \emptyset$
\For{$\{M_C,P_C,S_C\}$ in $\mathcal{W}$ in chronological order}
    \State $\mathbf{x}_{\text{raw}} \gets M_C$ \Comment{Windowed sensor measurements}
    \State $\mathbf{x}_{\text{filt}} \gets \mathsf{LowPass}(\mathbf{x}_{\text{raw}}, f_c)$
    \State $\mathbf{x}_{\text{ds}} \gets \mathsf{Downsample}(\mathbf{x}_{\text{filt}}, d)$
    \State $\mathbf{s} \gets \mathsf{Enc}(\mathbf{x}_{\text{ds}}, P_C, S_C)$ \Comment{Encoded sensor string}
    \State $\mathbf{t} \gets \Pi(\mathbf{s})$ \Comment{Token sequence}
    \State $\mathcal{T}^+ \gets \mathcal{T}^+ \cup \{(\mathbf{t}, P_C, S_C)\}$
\EndFor
\State \Return $\mathcal{T}^+$
\end{algorithmic}
\end{algorithm}

\subsubsection{LLM-Based Behavior Prediction Pipeline}

The LLM-based behavior prediction pipeline takes the tokenized sensor string $\mathcal{T}^+$ as its primary input. In addition, the pipeline receives a prompt $\rho$ that defines the forecasting task and relevant operational context, along with model control parameters such as the temperature $\tau$ and top-$p$ sampling threshold $p_{\text{top}}$. The tokenized string, prompt, and parameters are jointly encoded to produce an input representation compatible with the LLM ensemble.

The ensemble consists of multiple LLM instances configured for time-series forecasting on tokenized process data. Each model in the ensemble generates a raw prediction string corresponding to the anticipated sensor values for the next process state $P_{c+1}$. These raw prediction strings are decoded into numerical signal predictions, forming a prediction matrix where each column corresponds to an ensemble member and each row represents a forecasted time step or measurement channel.

To produce a robust final forecast, the system computes the median across ensemble predictions for each time step and measurement dimension. This median-based aggregation mitigates the influence of outlier forecasts and provides a stable estimate of the future process state. The result is a statistical point estimate of the signal values expected at $P_{c+1}$.

Finally, the predicted values are evaluated against expected operational ranges and thresholds to determine the health status of the physical process. If the forecasted measurements indicate potential degradation, anomaly, or failure, the system flags the process for further analysis or intervention. This health assessment capability ensures that the LLM-based prediction engine not only forecasts future states but also contributes to proactive performance assurance within the Industry 4.0 environment.

\begin{algorithm}[t]
\caption{End-to-End DDD-GenDT with DDDAS Feedback}
\label{alg:ddd-gendt-dddas}
\begin{algorithmic}[1]
\Require PTOG database $DB$ with measurements $M_{c}^{(r)}$; epoch set $\mathcal{E}_p \subseteq \mathcal{R}\times\mathcal{P}$; window length $L_w$; cutoff $f_c$; downsampling factor $d$; encoder $\mathsf{Enc}(\cdot)$; tokenizer $\Pi(\cdot)$; prompt $\rho$; parameters $(\tau, p_{\text{top}})$; LLM $\mathcal{M}_{LLMfx}$; reconstructions $n$; thresholds $(T_{\text{low}},T_{\text{high}},T_{\text{health}})$
\Ensure Forecasts $\{\hat{Y}_c\}$ and control actions $\{U(Q_c)\}$ at epoch points
\For{each $(R_{r_c}, P_c) \in \mathcal{E}_p$ in chronological order}
    \State $S_C \gets$ sensor assigned to $P_c$
    \State $\mathcal{X}_{\text{hist}} \gets [\,]$ \Comment{Observation window for $P_c$ across runs}
    \State $r \gets r_c$
    \While{$|\mathcal{X}_{\text{hist}}| < L_w$ \textbf{and} $r \ge 1$}
        \If{$DB$ contains $\mathbf{x}_{k}^{r}$ for $k=\mathcal{P}_c$}
            \State Append $\mathbf{x}_{k}^{r}$ to $\mathcal{X}_{\text{hist}}$
        \EndIf
        \State $r \gets r-1$
    \EndWhile
    \State $\mathbf{x}_{\text{filt}} \gets F_{f_c}(\mathcal{X}_{\text{hist}})$
    \State $\mathbf{x}_{\text{ds}} \gets D_d(\mathbf{x}_{\text{filt}})$
    \State $\mathbf{s} \gets \mathsf{Enc}(\mathbf{x}_{\text{ds}}, P_c, S_C)$
    \State $\mathbf{t} \gets \Pi(\mathbf{s})$
    \State $DT_c \gets \mathcal{M}_{LLMfx}(\mathbf{t}, \rho, p_{\text{top}}, \tau, L_w)$
    \State $DT_c^{out} \gets$ empty matrix of size $0 \times L_w$
    \For{$i \gets 1$ to $n$}
        \State $\hat{\mathbf{y}}_i \gets DT_c.\textit{output}()$ \Comment{$\hat{\mathbf{y}}_i \in \mathbb{R}^{L_w}$}
        \State Append $\hat{\mathbf{y}}_i$ as a new row to $DT_c^{out}$
    \EndFor
    \State $\hat{Y}_c \gets \text{median}(DT_c^{out})$ \Comment{Elementwise median across rows}
    \State Obtain $O_c$ from $DB$ for $(R_{r_c},P_c)$ if available
    \If{$O_c$ available}
        \State $Q_c \gets \mathrm{RMSE}(O_c,\hat{Y}_c)$
        \State $U(Q_c) \gets 
        \begin{cases}
            \text{Continue} & Q_c < T_{\text{low}}\\
            \text{Warning} & T_{\text{low}} \le Q_c \le T_{\text{high}}\\
            \text{Stop} & Q_c > T_{\text{high}}
        \end{cases}$
    \Else
        \State $Q_c \gets$ not evaluated,\quad $U(Q_c) \gets$ no action
    \EndIf
    \State Store $\hat{Y}_c$, $Q_c$, and $U(Q_c)$; update $DB$ with new measurements when received
\EndFor
\State \Return $\{\hat{Y}_c\}$, $\{U(Q_c)\}$
\end{algorithmic}
\end{algorithm}

\subsection{Linking DDD-GenDT to DDDAS}

The Dynamic Data-Driven Generative Digital Twin (DDD-GenDT) framework operationalizes the core concepts of the Dynamic Data-Driven Applications Systems (DDDAS) paradigm through its integration of the PTOG, observation window extraction, and LLM-based behavior prediction engine. DDDAS defines a closed feedback loop in which real-time instrumentation data is assimilated into executing models, and the updated model states are used to influence the instrumentation process. DDD-GenDT instantiates this paradigm in the context of industrial process monitoring and forecasting.

In the physical space, measurements from multiple prior runs from the PT, denoted as $\{M_{C-n}, \cdots M_{C-1}, M_{C}\}$, are collected from sensors attached to process states. These measurements represent mechanical changes over time for corresponding process steps \cite{xu2023review}. The PTOG captures and organizes these data streams across runs, preserving both temporal and structural relationships between process states and sensor readings. This representation enables the assimilation of historical and current measurement data into the forecasting process, fulfilling the DDDAS requirement for continuous data integration. In the digital space, the PTOG serves as the model’s data foundation. At specific points along the process timeline, called Epoch Points, the framework triggers future state generation. These points, indicated by colored circles in the figure, correspond to well-defined process transitions where both sufficient data context and predictive relevance are available. The LLM-based behavior prediction engine uses the observation windows that include measurements from the current and past runs for the same process state to generate the next process state $P_{c+1}$. This design mirrors DDDAS’s model execution step, where the model dynamically evolves based on the latest assimilated data.

The prediction results at each Epoch Point provide not only the expected sensor signals for $P_{c+1}$ but also health assessments for the physical process. These outputs are evaluated against operational thresholds and used to determine whether adjustments in control inputs are required. This process embodies DDDAS’s instrumentation control function, where model outputs influence the acquisition strategy, sensor configurations, or actuator commands in the physical system.

The continuous interaction between physical and digital spaces in DDD-GenDT reflects the computational integration aspect of DDDAS. The architecture ensures that the data acquisition, predictive modeling, and control decision-making processes operate in a coordinated cycle. The physical measurements feed the digital model through PTOG assimilation, the model forecasts future states and health conditions, and the predictions inform physical space control actions. This alignment enables adaptive and proactive management of industrial processes.

By combining GenAI-based forecasting with structured historical data representation and closed-loop decision-making, DDD-GenDT extends the DDDAS paradigm to scenarios where process variability, mechanical degradation, and complex temporal dependencies must be addressed. The framework ensures that predictions are informed by both short-term state dynamics and long-term operational patterns, allowing it to adapt as process conditions evolve. The linkage between DDD-GenDT and DDDAS thus establishes a robust methodology for real-time process optimization and health assurance in Industry 4.0 environments.

\begin{table}[tbp!]
\centering
\caption{Notations in This Article}
\label{tab:notations}
\resizebox{\columnwidth}{!}{%
\begin{tabular}{cl}
\hline\hline
\textbf{Notation} & \textbf{Meaning} \\
\hline
$\mathbb{N}^+$ & Positive integers $\{x \in \mathbb{N}\,|\,x>0\}$ \\
$\mathcal{R}$ & Set of runs $\{R_1,\dots,R_{|\mathcal{R}|}\}$ \\
$\mathcal{P}$ & Set of process states $\{P_1,\dots,P_{|\mathcal{P}|}\}$ \\
$\mathcal{S}$ & Set of sensors \\
$\mathcal{G}=(\mathcal{V},\mathcal{E})$ & PTOG graph with vertices and edges \\
$\mathcal{V}$ & Vertex set of PTOG, nodes $(R_r,P_c)$ \\
$\mathcal{E}$ & Edge set, intra–run succession and cross–run alignment \\
$\mathcal{E}_p$ & Set of Epoch Points that trigger forecasting \\
$R_r$ & The $r$-th run \\
$P_c$ & The $c$-th process state \\
$S_C$ & Sensor assigned to state $P_c$ \\
$M_{c}^{(r)}$ & Measurement of $S_C$ at $(R_r,P_c)$ \\
$\mathbf{x}_{k}^{r}$ & Time series for state index $k$ in run $r$ \\
$\mathcal{X}_{\text{hist}}$ & Observation window collected across runs \\
$L_w$ & Observation window length \\
$F_{f_c}(\cdot)$ & Low–pass filter with cutoff $f_c$ \\
$D_d(\cdot)$ & Downsampling operator with factor $d$ \\
$\mathsf{Enc}(\cdot)$ & Encoder to structured sensor string \\
$\Pi(\cdot)$ & Tokenizer mapping strings to tokens \\
$\mathcal{T}^+$ & Tokenized dataset after preprocessing \\
$\rho$ & Forecast prompt for the LLM \\
$\tau$ & Temperature parameter of the LLM \\
$p_{\text{top}}$ & Top–$p$ parameter of the LLM \\
$\mathcal{M}_{LLMfx}(\cdot)$ & LLM function used for forecasting \\
$DT_c$ & Current–state $DT$ predictor $:=\mathcal{M}_{LLMfx}(\cdot)$ \\
$DT_c^{out}\in\mathbb{R}^{n\times L_w}$ & Matrix of $n$ forecast attempts over $L_w$ steps \\
$\hat{\mathbf{y}}_i$ & Forecast vector from the $i$-th attempt \\
$\mathbf{\hat{t}}_j$ & Column vector at step $t_j$ in $DT_c^{out}$ \\
$\hat{Y}_c$ & Median–aggregated forecast for state $P_c$ or $P_{c+1}$ \\
$O_c$ & Observation vector for current state $P_c$ \\
$Q_c$ & Quantity of interest, e.g., $\mathrm{RMSE}(O_c,\hat{Y}_c)$ \\
$e_c^{\mathrm{RMSE}}$ & RMSE at state $P_c$ \\
$e_{\mathrm{cum}}$ & Cumulative RMSE over states \\
$U(\cdot)$ & Control input decision function \\
$R(\cdot)$ & Reward or health evaluation function \\
$T_{\text{low}},T_{\text{high}}$ & Decision thresholds for $U(\cdot)$ \\
$T_{\text{health}}$ & Production–level health threshold \\
$f_c$ & Low–pass cutoff frequency \\
$d$ & Downsampling factor \\
$DT$ & Dynamic data–driven digital twin (generic symbol) \\
$M$ & Physical measurement (generic symbol) \\
$C$ & Current measurement index in a run \\
$Y$ & Ground truth signal \\
$\hat{Y}$ & Predicted signal \\
$\mathbf{t}$ & Vector collecting identical time indices across attempts \\
$T$ & Threshold (generic) \\
$r$ & Record or run index \\
$t\in\mathbb{N}^+$ & Discrete time index of observation \\
$\hat{t}\in\mathbb{N}^+$ & Discrete time index of $DT$ output \\
$c\in\mathbb{N}^+$ & Process state index \\
$k\in\mathbb{N}^+$ & State index used in PTOG extraction \\
$i\in\mathbb{N}^+$ & Attempt or generic index \\
$n\in\mathbb{N}^+$ & Number of reconstruction attempts \\
$j\in\mathbb{N}^+$ & Position index within the window \\
$m\in\mathbb{N}^+$ & Number of historical records retrieved \\
\hline\hline
\end{tabular}%
}
\end{table}

Algorithm \ref{alg:ddd-gendt-dddas} provides the procedural view of the DDD-GenDT framework, encapsulating the end-to-end workflow from PTOG-based measurement retrieval to LLM-driven forecasting and DDDAS-style feedback control. At each Epoch Point, the algorithm retrieves the relevant observation window from the PTOG, applies the data preprocessing pipeline, and generates multi-sample forecasts for the next process state via an LLM ensemble. These predictions are aggregated using the median operator to produce point estimates, compared against real measurements when available, and evaluated through a tiered health assessment policy to determine corresponding control actions. This operational pipeline directly maps to the DDDAS feedback loop, where continuous measurement assimilation, adaptive model execution, and instrumentation control are executed in a tightly coupled cycle. The mathematical formulation of each stage is provided in Section \ref{sec:formal_model}.

\subsection{Formal Model of the DDD-GenDT Framework} \label{sec:formal_model}

The DDD-GenDT framework implements the core principles of the Dynamic Data-Driven Applications Systems (DDDAS) paradigm for industrial process forecasting and health assurance. The framework comprises three main stages: (1) physical measurement organization through the PTOG, (2) data preprocessing and mapping to the LLM-based behavior prediction engine, and (3) feedback-based decision control for physical process adaptation. This section presents a formal description of the end-to-end process.

\subsubsection{Physical Measurement Representation via PTOG}

In the physical space, let $\mathcal{R} = \{R_1, R_2, \dots, R_{|\mathcal{R}|}\}$ be the set of runs executed in chronological order, and let $\mathcal{P} = \{P_1, P_2, \dots, P_{|\mathcal{P}|}\}$ denote the set of process states. Each process state $P_c \in \mathcal{P}$ is monitored by an associated sensor $S_C \in \mathcal{S}$. For run $R_r$ and process state $P_c$, the measurement from $S_C$ is denoted $M_{c}^{(r)} \in \mathbb{R}^{T_s}$, where $T_s$ is the number of samples in the state. The PTOG encodes this mapping as
\begin{equation}
\mathcal{G} = (\mathcal{V}, \mathcal{E}), \quad \mathcal{V} = \{(R_r, P_c)\}, \quad \mathcal{E} \subseteq \mathcal{V} \times \mathcal{V}
\end{equation}
Edges in $\mathcal{E}$ capture temporal succession within runs and cross-run state alignment for identical $P_c$ across different runs.

\subsubsection{Observation Window Extraction at Epoch Points}

Let $L_w$ be the observation window length. For the current run $R_{r_c}$ and process state $P_c$, the observation window $\mathcal{X}_{\text{hist}}$ is defined as the concatenation of $L_w$ most recent measurements for $P_c$, collected across the current and preceding runs:
\begin{equation} \label{equ:input_data}
\mathcal{X}_{\text{hist}} = [\mathbf{x}_{k}^{r_{c-L_w}}, \mathbf{x}_{k}^{r_{c-L_w+1}}, \dots, \mathbf{x}_{k}^{r_{c-1}}]_{k = \mathcal{P}_c}
\end{equation}
Here $\mathbf{x}_{k}^{r}$ denotes the time series measurement vector for process state index $k$ in run $r$. The set of Epoch Points $\mathcal{E}_p \subseteq \mathcal{R} \times \mathcal{P}$ defines the locations along the process timeline at which a forecast for $P_{c+1}$ is triggered. Each Epoch Point is chosen such that both adequate historical context and predictive relevance are available.

\begin{figure*}[t!]
\includegraphics[width=0.9\textwidth]{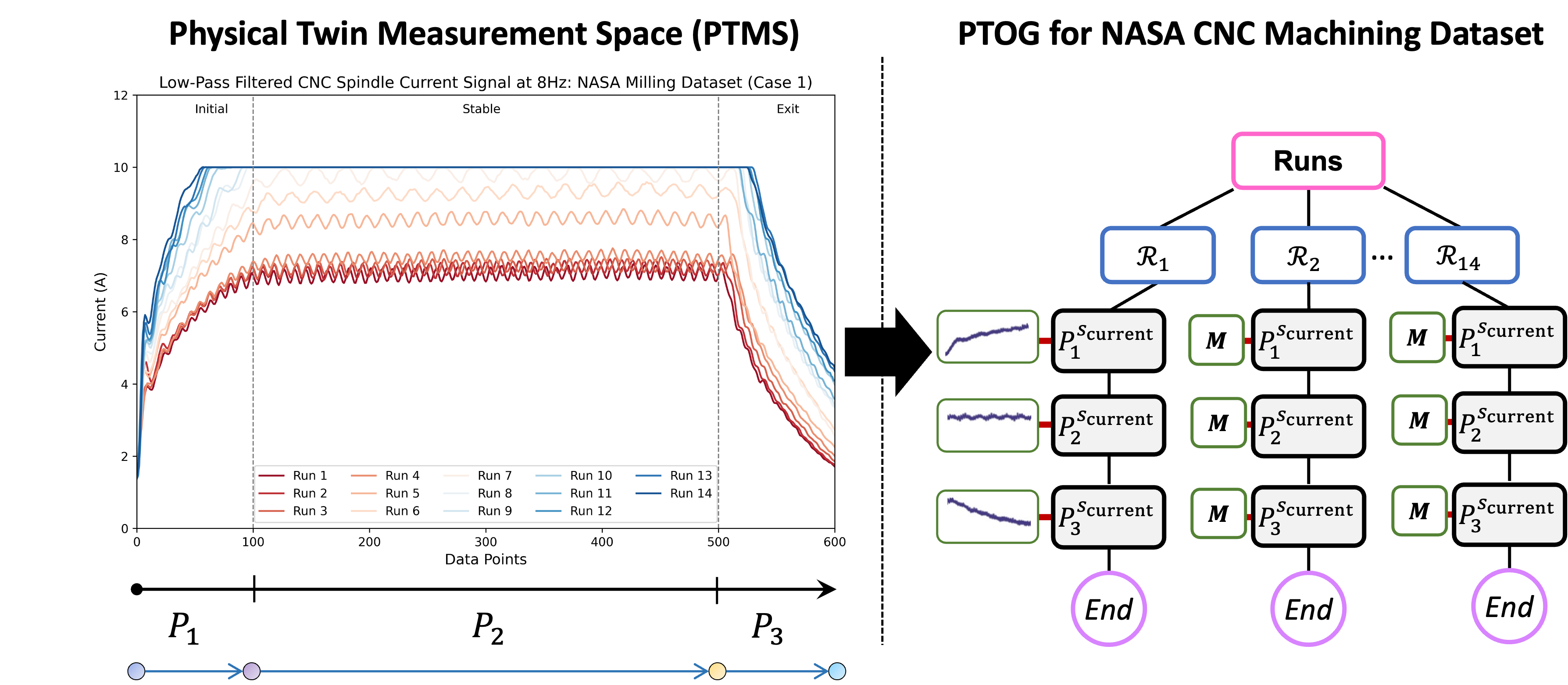}
\caption{Mapping the Physical Twin Measurement Space (PTMS) to the Physical Twin Observation Graph (PTOG) for the NASA CNC milling dataset. Left: low-pass filtered spindle current signals (8~Hz cutoff) for 14 runs, segmented into three process states $P_1$, $P_2$, and $P_3$. Right: PTOG representation preserving temporal ordering and run structure for downstream prediction in the DDD-GenDT framework.}
\label{fig:nasa_ptms2ptog}
\end{figure*} 

\subsubsection{Data Preprocessing Pipeline}

Each observation window $\mathcal{X}_{\text{hist}}$ undergoes a preprocessing sequence before LLM-based prediction. The sequence is defined as:
\begin{enumerate}
    \item \textbf{Low-pass filtering:} $F_{f_c}(\mathcal{X}_{\text{hist}})$ removes high-frequency noise using cutoff frequency $f_c$.
    \item \textbf{Downsampling:} $D_d(\cdot)$ reduces sample density by keeping every $d$-th sample to improve computational efficiency.
    \item \textbf{Encoding:} $\mathsf{Enc}(\cdot)$ transforms the numerical series into a structured sensor string preserving temporal order and state identity.
    \item \textbf{Tokenization:} $\Pi(\cdot)$ maps the encoded string into a discrete token sequence $\mathbf{t} \in \mathcal{V}^*$ for LLM compatibility.
\end{enumerate}
The composition of these operators is expressed as:
\begin{equation}
\mathcal{T}^+ = \Pi \circ \mathsf{Enc} \circ D_d \circ F_{f_c}(\mathcal{X}_{\text{hist}})
\end{equation}
where $\mathcal{T}^+$ is the tokenized dataset ready for LLM-based forecasting.

\subsubsection{LLM-Based Behavior Prediction}

The LLM-based behavior prediction engine aims to map the current physical state $P_c$ to a forecasted future state $P_{c+1}$. Let $\rho$ denote the forecast prompt, $\tau$ the temperature, and $p_{\text{top}}$ the top-$p$ threshold. The tokenized sequence $\mathcal{T}^+$, prompt $\rho$, and parameters $(\tau, p_{\text{top}})$ are jointly encoded for inference:
\begin{equation}
DT_c := \mathcal{M}_{LLMfx}(\mathcal{T}^+, \rho, p_{\text{top}}, \tau, L_w)
\end{equation}
where $\mathcal{M}_{LLMfx}(\cdot)$ denotes the LLM function.

To reduce variability in LLM outputs, the model is executed $n$ times. Each execution returns a forecast vector
\begin{equation}
\hat{\mathbf{y}}_i = \begin{bmatrix} \hat{y}_i(t_1) & \hat{y}_i(t_2) & \dots & \hat{y}_i(t_{L_w}) \end{bmatrix}
\end{equation}
The $n$ attempts are aggregated into the matrix:
\begin{equation}
DT_{c}^{out} = \begin{bmatrix}
\hat{\mathbf{y}}_1 \\
\hat{\mathbf{y}}_2 \\
\vdots \\
\hat{\mathbf{y}}_n
\end{bmatrix} \in \mathbb{R}^{n \times L_w}
\end{equation}

Let $\mathbf{\hat{t}}_j$ be the $j$-th column of $DT_c^{out}$:
\begin{equation}\label{equ:D_out2vector}
DT_{c}^{out} = 
\begin{bmatrix}
    \vert & \vert &  & \vert\\ 
    \mathbf{\hat{t}}_1 & \mathbf{\hat{t}}_2  & \dots & \mathbf{\hat{t}}_{L_w}  \\
    \vert & \vert &  & \vert
\end{bmatrix}
\end{equation}
\begin{equation}
\mathbf{\hat{t}}_j = \begin{bmatrix} \hat{y}_1(t_j) & \hat{y}_2(t_j) & \dots & \hat{y}_n(t_j) \end{bmatrix}^T
\end{equation}
The final point estimate is computed via the median:
\begin{equation}
\hat{Y}_c = \text{median}(DT_c^{out}) = \begin{bmatrix} \widetilde{\mathbf{t}}_1 & \widetilde{\mathbf{t}}_2 & \dots & \widetilde{\mathbf{t}}_{L_w} \end{bmatrix}
\end{equation}

\subsubsection{Feedback Control and Health Assessment}

Let $O_c$ be the observation vector for the current state:
\begin{equation}
O_c = \begin{bmatrix} o_{c,1} & o_{c,2} & \dots & o_{c,L_w} \end{bmatrix}, \quad o \in \mathbb{R}
\end{equation}
The quantities of interest $Q_c$ are computed as the RMSE between $O_c$ and $\hat{Y}_c$:
\begin{equation}
Q_c = e_c^{\text{RMSE}} = \text{RMSE}(O_c, \hat{Y}_c)
\end{equation}
The control decision $U(Q_c)$ follows a three-level policy:
\begin{equation}
U(Q_c) = 
\begin{cases} 
\text{Continue} & Q_c < T_{low} \\
\text{Warning} & T_{low} \leq Q_c \leq T_{high} \\
\text{Stop} & Q_c > T_{high}
\end{cases}
\end{equation}

\subsubsection{Production-Level Health Evaluation}

Over the course of production, the cumulative RMSE is:
\begin{equation}
e_{\text{cum}} = \sum_{i=1}^{c} e_{i}^{\text{RMSE}}
\end{equation}
The reward function $R(e_{\text{cum}})$ assesses the overall product quality:
\begin{equation}
R(e_{\text{cum}}) = 
\begin{cases} 
\text{Fail} & e_{\text{cum}} > T_{health} \\
\text{Pass} & e_{\text{cum}} \leq T_{health}
\end{cases}
\end{equation}
The thresholds $T_{low}$, $T_{high}$, and $T_{health}$ are tunable system parameters selected according to process-specific tolerances. This feedback path aligns with the instrumentation control loop in the DDDAS architecture, where forecast outputs directly inform real-time process adjustments.

\begin{table*}[h!]
\centering
\caption{Prediction Results of Spindle Current}
\begin{tabular}{cccccccccc}
\hline
                     & \multicolumn{1}{l}{}          & \multicolumn{8}{c}{Spindle Current Reconstruction (A)}                                                                                                                                                                                                             \\ \cline{3-10} 
Runs                 & Flank Wear (mm)               & \multicolumn{2}{c}{GPT 3.5 Turbo} & \multicolumn{2}{c}{GPT-4} & \multicolumn{2}{c}{\begin{tabular}[c]{@{}c@{}}1D CNN AE \\ (Training with Run 1-4)\end{tabular}} & \multicolumn{2}{c}{\begin{tabular}[c]{@{}c@{}}1D CNN AE\\ (Training with Run 3-4)\end{tabular}} \\
\multicolumn{1}{c}{} & \multicolumn{1}{c}{}          & $\mathbf{Err}_{avg}$        & $\mathbf{Err}_{std}$        & $\mathbf{Err}_{avg}$    & $\mathbf{Err}_{std}$    & $\mathbf{Err}_{avg}$                                        & $\mathbf{Err}_{std}$                                       & $\mathbf{Err}_{avg}$                                      & $\mathbf{Err}_{std}$                                       \\ \hline
1                    & 0                             & -               & -               & -           & -           & -                                               & -                                              & -                                              & -                                              \\
2                    & 0.0367 (Linear interpolation) & -               & -               & -           & -           & -                                               & -                                              & -                                              & -                                              \\
3                    & 0.0734 (Linear interpolation) & -               & -               & -           & -           & -                                               & -                                              & -                                              & -                                              \\
4                    & 0.11                          & -               & -               & -           & -           & -                                               & -                                              & -                                              & -                                              \\
5                    & 0.155 (Linear interpolation)  & 2.791           & 0.718           & 0.951       & 0.244       & 0.162                                           & 0.224                                          & 0.419                                          & 0.328                                          \\
6                    & 0.2                           & 1.832           & 0.867           & 0.812       & 0.504       & 0.173                                           & 0.246                                          & 0.449                                          & 0.362                                          \\
7                    & 0.24                          & 1.304           & 1.01            & 0.608       & 0.425       & 0.204                                           & 0.286                                          & 0.53                                           & 0.409                                          \\
8                    & 0.29                          & 1.002           & 1.127           & 0.553       & 0.661       & 0.172                                           & 0.27                                           & 0.464                                          & 0.398                                          \\
9                    & 0.28                          & 0.86            & 0.861           & 0.34        & 0.578       & 0.173                                           & 0.269                                          & 0.462                                          & 0.396                                          \\
10                   & 0.29                          & 0.52            & 0.848           & 0.385       & 0.685       & 0.174                                           & 0.277                                          & 0.468                                          & 0.405                                          \\
11                   & 0.38                          & 0.4             & 0.68            & 0.27        & 0.458       & 0.165                                           & 0.270                                          & 0.442                                          & 0.407                                          \\
12                   & 0.4                           & 0.338           & 0.651           & 0.352       & 0.679       & 0.102                                           & 0.187                                          & 0.27                                           & 0.315                                          \\
13                   & 0.43                          & 0.281           & 0.55            & 0.283       & 0.598       & 0.118                                           & 0.214                                          & 0.317                                          & 0.352                                          \\
14                   & 0.45                          & 0.257           & 0.516           & 0.243       & 0.495       & 0.119                                           & 0.213                                          & 0.314                                          & 0.349                                          \\ \hline
\end{tabular}
\label{tab:exp_result}
\end{table*}

\section{Experiment and Result Analysis}\label{sec:exp}

This section presents the evaluations for the proposed DDD-GenDT Framework, wherein we evaluate the proposed DDDGen-DT framework, and its components including the Physical Twin Observation Graph, the Observation Window extraction process, and the LLM based Behavior Prediction Engine as described in Section \ref{sec:ref_architecture}, quantifying the framework's ability to predict the PT behavior with little to no training data, and the ability to autonomically modify behavior as the PT ages, reflecting DT-aging.

\subsection{Experimental Setup}

This subsection decribes the Experimental Setup for the proposed evaluations.

\subsubsection{DDD-GenDT Architecture Implementation with Generative Pre-Trained Transformer (GPT) LLMs at the prediction core}
The proposed DDD-GenDT framework from the Section \ref{sec:ref_architecture}, were implemented in Python 3 with the official OpenAI API library allowing the LLM-Based Behavior Prediction Pipeline described in Algorithm \autoref{alg:ddd-gendt-dddas} to use GPT-3.5 and GPT-4 ensembles for the PT behavior prediction. The Data Preprocessing Pipeline described in Algorithm \autoref{alg:data_preprocessing_pipeline} was configured to the following settings: 
\begin{itemize}
    \item Number of Observation Window: 4 Process States from prior runs.
    \item Low Pass Filter Cutoff ($f_c$): 8Hz via a Butterworth Filter.
    \item Downsampling Factor ($d$): 20Hz.
\end{itemize}

The following prompt and temperature parameters were used in the LLM-based Behavior Prediction Pipeline:

\begin{itemize}
    \item Prompt ($\rho$) Definition:\begin{quote}
\texttt{
You are a helpful assistant who performs time series predictions.
The user will provide a sequence, and you will predict the remaining sequence.
The sequence is represented by decimal strings separated by commas.
Please continue the following sequence without producing any additional text.
Do not say anything like 'the next terms in the sequence are', just return the numbers.
Sequence: <ENCODED TOKENIZED STRING>.}
\end{quote}
    \item Temperature parameter for the LLM ($\tau$): 
    \begin{itemize}
        \item GPT-3.5-Turbo: 0.7.
        \item GPT-4: 1.0.
    \end{itemize}
    \item Top-p parameter for the LLM ($p_{\text{top}}$): 1.0.
    \item Number of reconstruction attempts (ensemble size) ($n$): 10.
\end{itemize}

\subsubsection{NASA Computer Numerical Control (CNC) Machining Dataset}

The DDD-GenDT framework is evaluated on a Computer Numerical Control (CNC) machining processe, wherein a CNC machine cuts through different metal structures to manufacture metallic pieces, while an array of sensors monitors and collects data. The dataset, primarily collected to represent tool wear caused by the gradual degradation of the CNC machine's cutting edge, comprises of 14 cutting cases called runs with each run being of irregular intervals, capturing parameters like depth of cut, feed speed, cutting speed, and current drawn. The 14 cutting cases of this dataset, in combination with the tool wear, make this dataset an ideal candidate for evaluation of the DDD-GenDT framework, which is expected to model the CNC machining process with little to no training datasets, and effectively demonstrate DT-aging as the tool wear degrades the cutting edge.

For the scope of this work, the evaluation of the DDD-GenDT framework is restricted to one sensor feed, representing the spindle current drawn as a time series. This measurement is chosen due to its significance as an indicator of both process quality and system security. Different machining behaviors produce distinct patterns in the spindle power consumption curves, enabling analysis and insight into the machining process. Prior studies have demonstrated that spindle power consumption, observed via side-channel measurements, can be used to detect product quality during cutting and prevent cyber-physical attacks \cite{lin2023dt4i4}. Furthermore, tool wear directly affects spindle motor current, making it a valuable metric for assessing tool condition \cite{kim2002real, uekita2017tool}. Figure \ref{fig:nasa_ptms2ptog} shows all the 14 Runs from the NASA CNC milling dataset, mapped to their corresponding Process States ($P_c$) and onto the PTOG. The CNC milling process comprises three linear Process States, each mapped in the PTOG to the corresponding current signal reading, and Table \ref{tab:exp_result} details the flank wear measurements across different runs.

\subsection{Experimental Analysis}
In the remainder of this section, we present the experiments evaluating the proposed DDD-GenDT framework.

\subsubsection{Experiment 1: Evaluation of DDD-GenDT's LLM-Based Behavior Prediction Engine to Predict PT Behavior}
This experiment evaluates the capability of the proposed DDD-GenDT framework to predict the future behavior of a PT without requiring any training data, by leveraging the LLM-Based Behavior Prediction Engine in conjunction with the PTOG. The evaluation uses the NASA Milling Dataset to assess the prediction performance across different process states defined in the PTOG, and compares the results against baseline 1D CNN Autoencoder (AE) models that require substantial training datasets. The goal is to demonstrate the feasibility and effectiveness of zero-shot LLM-based prediction for PT behavior reconstruction in Industry 4.0 settings.

\begin{figure}[t!]
\centering
\includegraphics[width=\columnwidth]{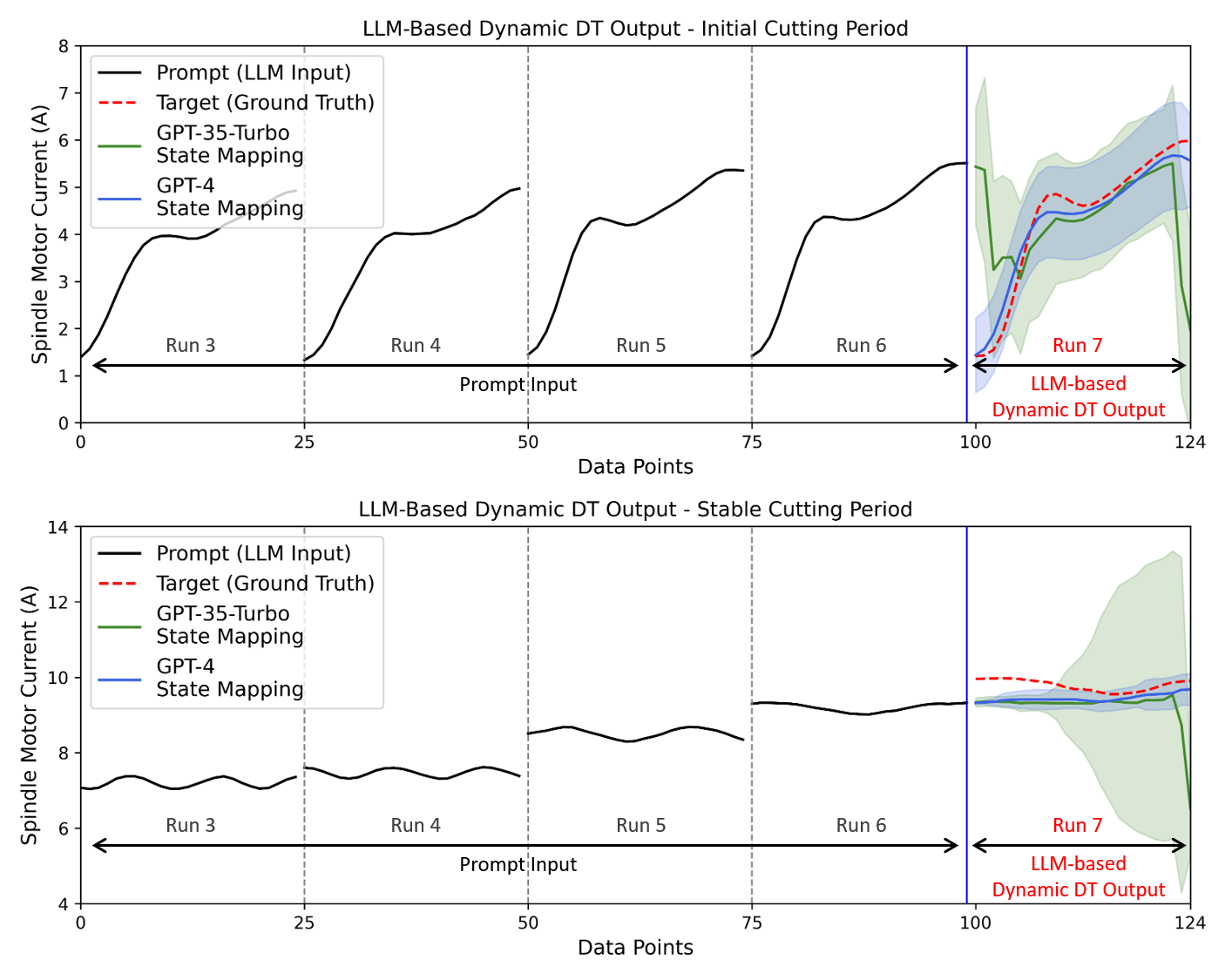}
\caption{Example results of mapping and predicting spindle motor current behavior. The black line is an ensemble time series input to LLM for mapping and predicting the behavior change in the same physical state.}
\label{fig:DDDT_Result}
\end{figure}

The evaluation follows the PTOG representation described in Figure \ref{fig:nasa_ptms2ptog}, where each machining run is decomposed into three process states: $P_1$ (initial cutting period), $P_2$ (stable cutting period), and $P_3$ (exit period). The PT Measurement Space (PTMS) in Figure \ref{fig:nasa_ptms2ptog} shows the low-pass filtered spindle motor current signals sampled at 8Hz, with distinct nonlinear and linear behaviors across different states. The PTOG maps these process states across multiple runs, forming structured temporal segments that serve as inputs to the LLM-based prediction process. In the DDD-GenDT framework, the observation window extracts a set of process state sequences from prior runs to serve as contextual input to the LLM ensemble. For example, when predicting $P_1$ in run 7, the observation window contains $P_1$ from runs 3-6. This sequence is encoded into a time-series representation and embedded into a prompt designed for zero-shot inference by the LLMs.

The LLM-Based Behavior Prediction Engine uses an ensemble of GPT-3.5 Turbo and GPT-4 models, each independently prompted with the observation window data. The outputs are aggregated by computing the median prediction across the ensemble for each model, along with a statistical uncertainty bound represented by $\pm$ one standard deviation (SD). This method captures the central tendency of the predictions while providing an estimate of prediction variability. Importantly, the framework does not rely on any prior training on the target dataset, in contrast to the baseline CNN models, which are trained on large subsets of the NASA Milling Dataset.

Figure \ref{fig:DDDT_Result} presents an example prediction for run 7. The upper panel shows predictions for $P_1$, a process state characterized by a nonlinear increase in spindle current due to initial cutting dynamics. Accurately predicting such nonlinear behavior is inherently challenging because small deviations in initial conditions can lead to significant trajectory variations. Nevertheless, both GPT-3.5 Turbo and GPT-4 achieve high accuracy in reconstructing the nonlinear profile, with GPT-4 exhibiting slightly lower prediction variance.The lower panel of Figure \ref{fig:DDDT_Result} shows predictions for $P_2$, the stable cutting period with a more linear spindle current pattern. While the complexity of prediction is lower in $P_2$, the results confirm that the LLM-based engine maintains consistency across different dynamical regimes.

\begin{figure}[t!]
\centering
\captionsetup{justification=centering}
\includegraphics[width=\columnwidth]{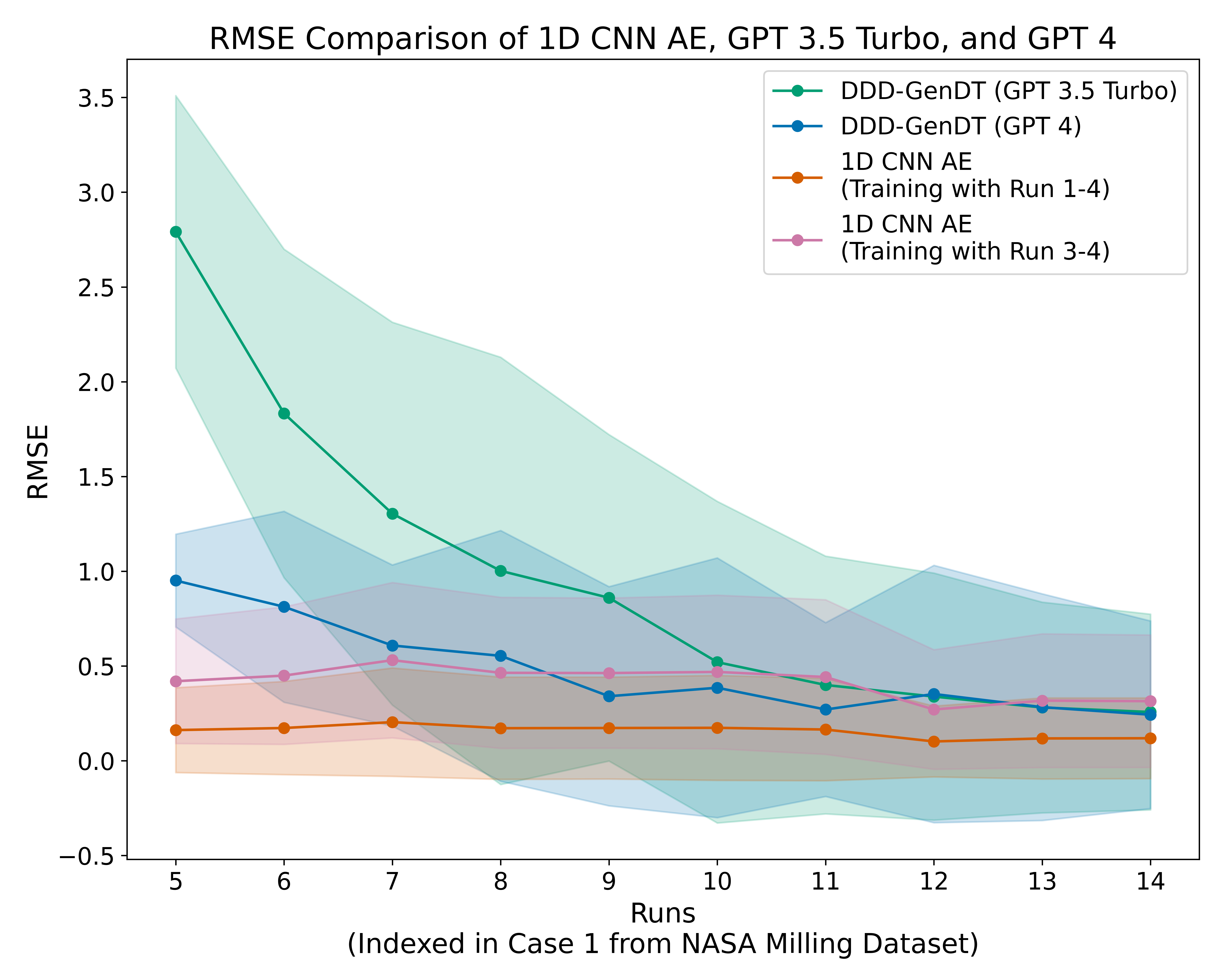}
\caption{Performance Comparison between DDD-GenDT and Autoencoder.}
\label{fig:Comparison}
\end{figure}

Table~\ref{tab:exp_result} quantifies the reconstruction accuracy in terms of average error ($Err_{avg}$) and standard deviation of error ($Err_{std}$) for both GPT-based predictions and the CNN baselines. The CNN baselines are trained on significantly larger datasets, either run 1-4 or runs 3-4, providing them with a clear advantage in terms of prior exposure. Despite this, the zero-shot GPT-4 model achieves competitive error levels, particularly in the mid-to-late runs, highlighting the potential of the approach for scenarios where training data is scarce or costly to obtain. The performance progression across runs is further illustrated in Figure \ref{fig:Comparison}, which plots the Root Mean Square Error (RMSE) for each method as the observation window slides through the PTOG. The RMSE curves for GPT-4 steadily decrease with increasing run index, approaching the performance of the CNN baselines in later runs, while GPT-3.5 Turbo exhibits a higher variance in early predictions but converges toward lower error values as more observation data becomes available.

The results demonstrate three key findings. First, the LLM-based prediction engine can accurately reconstruct PT behavior in a zero-shot setting without relying on pre-collected training data. This is critical for Industry 4.0 systems, where data acquisition may be expensive due to production costs or restricted due to intellectual property concerns. Second, the ability to predict nonlinear PT behaviors with high accuracy, as shown in $P_1$ predictions, indicates that LLMs can generalize temporal patterns from small observation windows to capture complex dynamic responses. This is a significant capability since traditional regression or neural network approaches typically require substantial training datasets to handle such variability. Third, although CNN baselines achieve lower RMSE in early runs due to their training advantage, the GPT-based models exhibit a clear learning trend over time through the DDD-GenDT observation mapping process. This suggests that the framework can leverage temporal correlations across runs to autonomically improve predictions as more PT operational history becomes available.

In summary, Experiment 1 confirms that the DDD-GenDT framework, through its LLM-Based Behavior Prediction Engine, can deliver competitive PT behavior prediction performance without any training phase, offering a viable alternative to traditional data-intensive modeling approaches. The combination of PTOG-based state mapping (Figure \ref{fig:nasa_ptms2ptog}), statistical ensemble prediction (Figure \ref{fig:DDDT_Result}), and robust RMSE trends (Figure \ref{fig:Comparison}) validates the effectiveness of the approach for real-world I4.0 systems.

\subsubsection{Experiment 2: Evaluation of DDD-GenDT's DT-Aging Capability}
The second experiment evaluates the DDD-GenDT framework’s ability to autonomously adapt its predictive behavior as the Physical Twin (PT) undergoes gradual aging. In the CNC milling process, PT aging is represented by the progressive wear of the cutting tool, which directly affects the spindle motor current profile. As the cutting edge becomes blunter, more torque is required to maintain material removal, leading to a gradual increase in the current draw over successive runs. The NASA milling dataset, as depicted in Figure~\ref{fig:nasa_ptms2ptog}, provides detailed spindle current measurements and associated tool wear readings, enabling the study of DT adaptation over time.

The evaluation procedure mirrors the methodology used in Experiment~1, with the observation window sliding across runs 5 to 14. For each run, the most recent process state measurements from the PTOG are provided as input to the LLM-based prediction engine, which then forecasts the next run’s behavior without any retraining or fine-tuning. Predictions are generated independently by the GPT-3.5 Turbo DT and GPT-4 DT, and their performance is compared using RMSE distributions.

\begin{figure}[t!]
\centering
\captionsetup{justification=centering}
\includegraphics[width=\columnwidth]{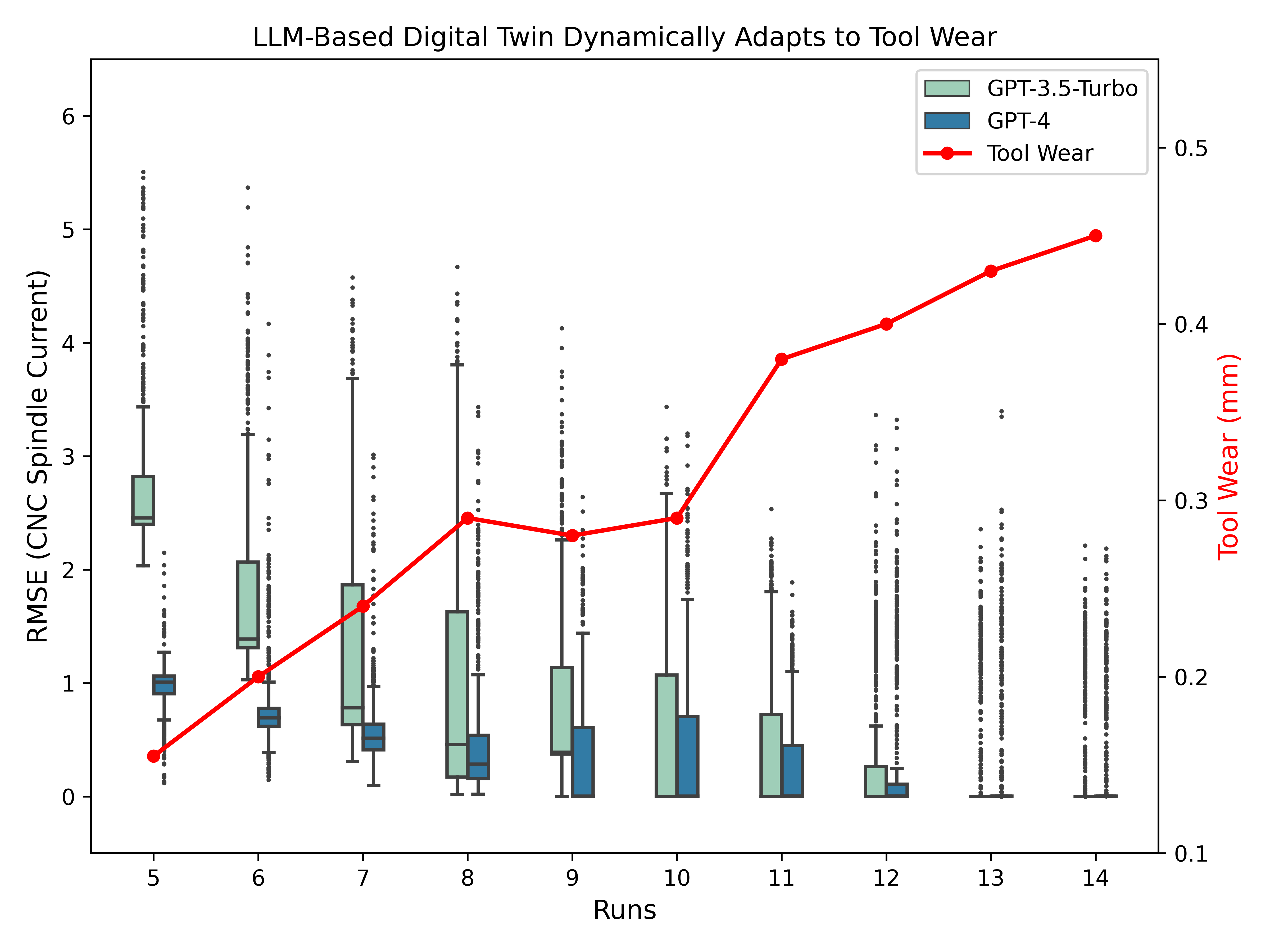}
\caption{Boxplot of LLM-Based Dynamic DT Performance}
\label{fig:Box_plot}
\end{figure}

The results, shown in Figure~\ref{fig:Box_plot}, present RMSE distributions as box plots for both LLM-based DTs, with GPT-3.5 Turbo represented in green and GPT-4 in blue. For each run, the median RMSE is indicated by the central line within each box, the box bounds correspond to the 25th and 75th percentiles, and whiskers extend to the minimum and maximum RMSE values. Outliers are plotted individually. Overlaid on the RMSE plots is a red line representing tool wear progression in millimeters, which shows a steady increase across the runs. This dual visualization enables direct assessment of the DT’s ability to track changes in PT dynamics while the tool physically degrades.

The results indicate that as the runs progress, both GPT-3.5 Turbo and GPT-4 DTs demonstrate a marked reduction in median RMSE and a narrowing of the interquartile range, signifying more accurate and consistent predictions. This trend persists despite the clear increase in tool wear, reflecting the framework’s capability to autonomously adapt to evolving physical system dynamics. GPT-4 consistently achieves lower RMSE values and reduced variability compared to GPT-3.5 Turbo, highlighting its superior ability to capture subtle nonlinear variations in the spindle current profile. The combined observation of improving accuracy, decreasing variance, and alignment with the tool wear trajectory confirms that the DDD-GenDT framework effectively embodies DT-aging behavior, maintaining high-fidelity predictions without the need for retraining, even as the PT progressively changes.

\section{Conclusion}\label{sec:conclusion}
This paper presents the Dynamic Data-Driven Generative Digital Twin (DDD-GenDT) framework, a novel architecture that integrates the Physical Twin Observation Graph (PTOG), Observation Window extraction, a Data Preprocessing Pipeline, and an LLM-based Behavior Prediction Engine with ensemble inference. The framework leverages generative AI capabilities to address fundamental challenges in digital twin construction for Industry 4.0, including the scarcity of training data, high integration complexity, and limited adaptability to evolving physical systems. By embedding DDD-GenDT within a Dynamic Data-Driven Application Systems (DDDAS) paradigm, the proposed approach enables autonomous adaptation of the digital twin to reflect physical twin aging, yielding accurate, context-aware behavior predictions without extensive prior training.

Evaluation on the NASA CNC milling dataset demonstrates that the LLM-based prediction engine achieves high accuracy in zero-shot settings, capturing both linear and nonlinear process state behaviors with competitive error margins when compared to data-intensive CNN autoencoders. Experimental results further confirm the DT-aging capability of the framework, as prediction accuracy improves over time in alignment with the wear progression of the physical system. These findings validate the ability of DDD-GenDT to construct resilient and adaptive digital twins that evolve in step with the physical system’s operational life.

The proposed architecture offers a generic and scalable methodology for digital twin creation across diverse application domains where data scarcity and system evolution present significant barriers. Future work will extend the framework to multi-modal sensing environments, explore adaptive prompt optimization for LLM ensembles, and integrate online learning mechanisms to enhance responsiveness to abrupt system changes.

\vspace*{-0.2cm}

\ifCLASSOPTIONcaptionsoff
  \newpage
\fi

\bibliography{refs.bib}
\bibliographystyle{IEEEtran}

\end{document}